# Joint Face Detection and Alignment using Multi-task Cascaded Convolutional Networks

Kaipeng Zhang, Zhanpeng Zhang, Zhifeng Li, *Senior Member, IEEE*, and Yu Qiao, *Senior Member, IEEE*

*Abstract*—Face detection and alignment in unconstrained environment are challenging due to various poses, illuminations and occlusions. Recent studies show that deep learning approaches can achieve impressive performance on these two tasks. In this paper, we propose a deep cascaded multi-task framework which exploits the inherent correlation between them to boost up their performance. In particular, our framework adopts a cascaded structure with three stages of carefully designed deep convolutional networks that predict face and landmark location in a coarse-to-fine manner. In addition, in the learning process, we propose a new online hard sample mining strategy that can improve the performance automatically without manual sample selection. Our method achieves superior accuracy over the state-of-the-art techniques on the challenging FDDB and WIDER FACE benchmark for face detection, and AFLW benchmark for face alignment, while keeps real time performance.

## I. Introduction

FACE detection and alignment are essential to many face applications, such as face recognition and facial expression analysis. However, the large visual variations of faces, such as occlusions, large pose variations and extreme lightings, impose great challenges for these tasks in real world applications.

The cascade face detector proposed by Viola and Jones [2] utilizes Haar-Like features and AdaBoost to train cascaded classifiers, which achieve good performance with real-time efficiency. However, quite a few works [1, 3, 4] indicate that this detector may degrade significantly in real-world applications with larger visual variations of human faces even with more advanced features and classifiers. Besides the cascade structure, [5, 6, 7] introduce deformable part models (DPM) for face detection and achieve remarkable performance. However, they need high computational expense and may usually require expensive annotation in the training stage. Recently, convolutional neural networks (CNNs) achieve remarkable progresses in a variety of computer vision tasks, such as image classification [9] and face recognition [10]. Inspired by the good performance of CNNs in computer vision tasks, some of the CNNs based face detection approaches have been proposed in recent years. Yang *et al*. [11] train deep convolution neural networks for facial attribute recognition to obtain high response in face regions which further yield candidate windows of faces. However, due to its complex CNN structure, this approach is time costly in practice. Li *et al*. [19] use cascaded CNNs for face detection, but it requires bounding box calibration from face detection with extra computational expense and ignores the inherent correlation between facial landmarks localization and bounding box regression.

Face alignment also attracts extensive interests. Regression-based methods [12, 13, 16] and template fitting approaches [14, 15, 7] are two popular categories. Recently, Zhang *et al*. [22] proposed to use facial attribute recognition as an auxiliary task to enhance face alignment performance using deep convolutional neural network.

However, most of the available face detection and face alignment methods ignore the inherent correlation between these two tasks. Though there exist several works attempt to jointly solve them, there are still limitations in these works. For example, Chen *et al*. [18] jointly conduct alignment and detection with random forest using features of pixel value difference. But, the handcraft features used limits its performance. Zhang *et al*. [20] use multi-task CNN to improve the accuracy of multi-view face detection, but the detection accuracy is limited by the initial detection windows produced by a weak face detector.

On the other hand, in the training process, mining hard samples in training is critical to strengthen the power of detector. However, traditional hard sample mining usually performs an offline manner, which significantly increases the manual operations. It is desirable to design an online hard sample mining method for face detection and alignment, which is adaptive to the current training process automatically.

In this paper, we propose a new framework to integrate these two tasks using unified cascaded CNNs by multi-task learning. The proposed CNNs consist of three stages. In the first stage, it produces candidate windows quickly through a shallow CNN. Then, it refines the windows to reject a large number of non-faces windows through a more complex CNN. Finally, it uses a more powerful CNN to refine the result and output facial landmarks positions. Thanks to this multi-task learning framework, the performance of the algorithm can be notably improved. The major contributions of this paper are summarized as follows: (1) We propose a new cascaded CNNs based framework for joint face detection and alignment, and carefully

K.-P. Zhang, Z.-F. Li and Y. Q. are with the Multimedia Laboratory, Shenzhen Institutes of Advanced Technology, Chinese Academy of Sciences, Shenzhen 518055, China. E-mail: kp.zhang@siat.ac.cn; zhifeng.li@siat.ac.cn; yu.qiao@siat.ac.cn

Z.-P. Zhang is with the Department of Information Engineering, The Chinese University of Hong Kong, Hong Kong. E-mail: zz013@ie.cuhk.edu.hk

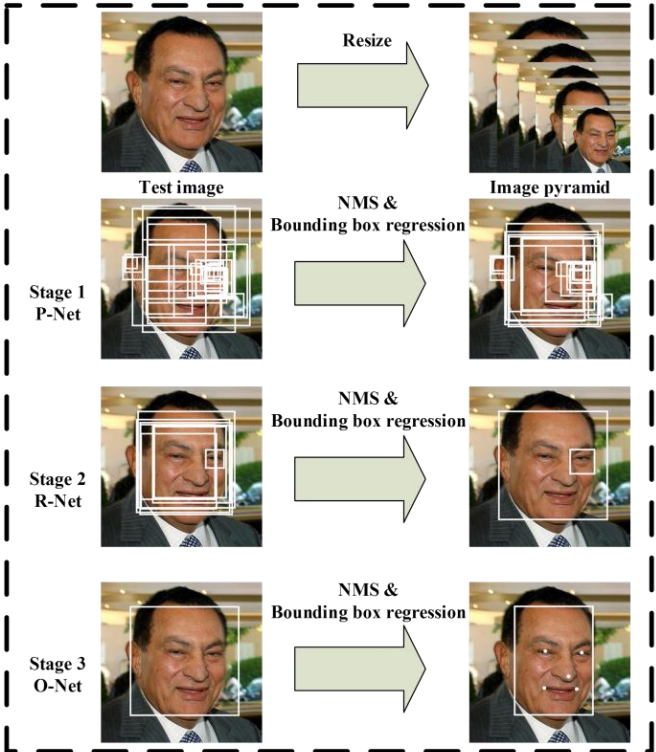

Fig. 1. Pipeline of our cascaded framework that includes three-stage multi-task deep convolutional networks. Firstly, candidate windows are produced through a fast Proposal Network (P-Net). After that, we refine these candidates in the next stage through a Refinement Network (R-Net). In the third stage, The Output Network (O-Net) produces final bounding box and facial landmarks position.

design lightweight CNN architecture for real time performance. (2) We propose an effective method to conduct online hard sample mining to improve the performance. (3) Extensive experiments are conducted on challenging benchmarks, to show the significant performance improvement of the proposed approach compared to the state-of-the-art techniques in both face detection and face alignment tasks.

## II. APPROACH

In this section, we will describe our approach towards joint face detection and alignment.

### A. Overall Framework

The overall pipeline of our approach is shown in Fig. 1. Given an image, we initially resize it to different scales to build an image pyramid, which is the input of the following three-stage cascaded framework:

**Stage 1**: We exploit a fully convolutional network[?], called Proposal Network (P-Net), to obtain the candidate windows and their bounding box regression vectors in a similar manner as [29]. Then we use the estimated bounding box regression vectors to calibrate the candidates. After that, we employ non-maximum suppression (NMS) to merge highly overlapped candidates.

**Stage 2**: all candidates are fed to another CNN, called Refine Network (R-Net), which further rejects a large number of false candidates, performs calibration with bounding box regression, and NMS candidate merge.

TABLE I
COMPARISON OF SPEED AND VALIDATION ACCURACY OF OUR CNNs AND PREVIOUS CNNs [19]

| Group  | CNN         | 300 Times Forward | Accuracy |
|--------|-------------|-------------------|----------|
| Group1 | 12-Net [19] | 0.038s            | 94.4%    |
| Group1 | P-Net       | 0.031s            | 94.6%    |
| Group2 | 24-Net [19] | 0.738s            | 95.1%    |
| Group2 | R-Net       | 0.458s            | 95.4%    |
| Group3 | 48-Net [19] | 3.577s            | 93.2%    |
| Group3 | O-Net       | 1.347s            | 95.4%    |

**Stage 3**: This stage is similar to the second stage, but in this stage we aim to describe the face in more details. In particular, the network will output five facial landmarks' positions.

### B. CNN Architectures

In [19], multiple CNNs have been designed for face detection. However, we noticed its performance might be limited by the following facts: (1) Some filters lack diversity of weights that may limit them to produce discriminative description. (2) Compared to other multi-class objection detection and classification tasks, face detection is a challenge binary classification task, so it may need less numbers of filters but more discrimination of them. To this end, we reduce the number of filters and change the 5×5 filter to a 3×3 filter to reduce the computing while increase the depth to get better performance. With these improvements, compared to the previous architecture in [19], we can get better performance with less runtime (the result is shown in Table 1. For fair comparison, we use the same data for both methods). Our CNN architectures are showed in Fig. 2.

### C. Training

We leverage three tasks to train our CNN detectors: face/non-face classification, bounding box regression, and facial landmark localization.

*1) Face classification:* The learning objective is formulated as a two-class classification problem. For each sample $x_i$, we use the cross-entropy loss:

$$L_i^{det} = -(y_i^{det} \log(p_i) + (1 - y_i^{det})(1 - \log(p_i))) \quad (1)$$

where $p_i$ is the probability produced by the network that indicates a sample being a face. The notation $y_i^{det} \in \{0,1\}$ denotes the ground-truth label.

*2) Bounding box regression:* For each candidate window, we predict the offset between it and the nearest ground truth (i.e., the bounding boxes' left top, height, and width). The learning objective is formulated as a regression problem, and we employ the Euclidean loss for each sample $x_i$:

$$L_i^{box} = \left\| \hat{y}_i^{box} - y_i^{box} \right\|_2^2 \quad (2)$$

where $\hat{y}_i^{box}$ regression target obtained from the network and $y_i^{box}$ is the ground-truth coordinate. There are four coordinates, including left top, height and width, and thus $y_i^{box} \in \mathbb{R}^4$.

*3) Facial landmark localization:* Similar to the bounding box regression task, facial landmark detection is formulated as a

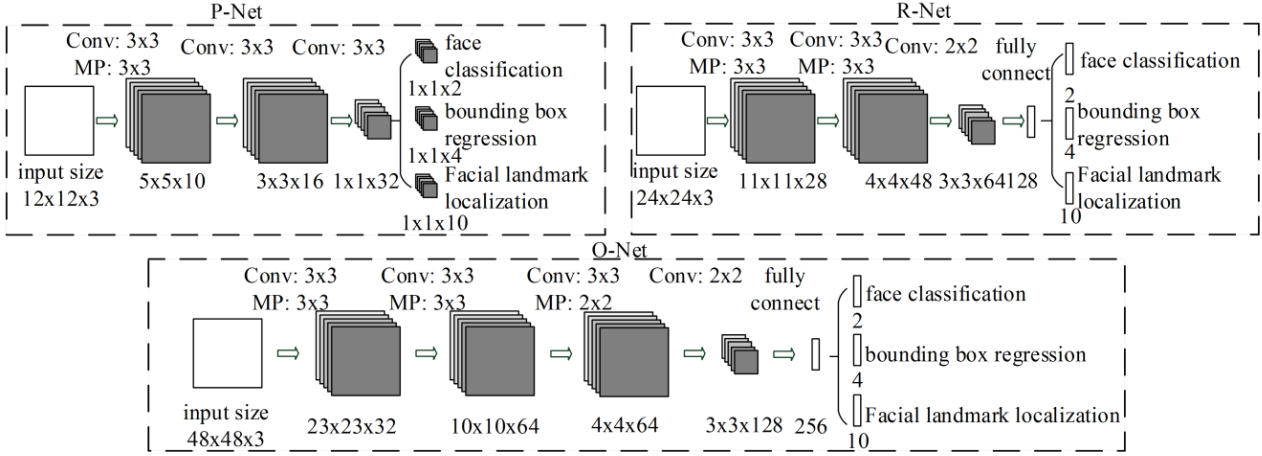

Fig. 2. The architectures of P-Net, R-Net, and O-Net, where "MP" means max pooling and "Conv" means convolution. The step size in convolution and pooling is 1 and 2, respectively.

regression problem and we minimize the Euclidean loss:

$$L_i^{landmark} = \left\| \hat{y}_i^{landmark} - y_i^{landmark} \right\|_2^2 \qquad (3)$$

where $\hat{y}_i^{landmark}$ is the facial landmark's coordinate obtained from the network and $y_i^{landmark}$ is the ground-truth coordinate. There are five facial landmarks, including left eye, right eye, nose, left mouth corner, and right mouth corner, and thus $y_i^{landmark} \in \mathbb{R}^{10}$.

*4) Multi-source training:* Since we employ different tasks in each CNNs, there are different types of training images in the learning process, such as face, non-face and partially aligned face. In this case, some of the loss functions (i.e., Eq. (1)-(3) ) are not used. For example, for the sample of background region, we only compute $L_i^{det}$, and the other two losses are set as 0. This can be implemented directly with a sample type indicator. Then the overall learning target can be formulated as:

$$\min \sum_{i=1}^{N} \sum_{j \in \{det, box, landmark\}} \alpha_j \beta_i^j L_i^j \qquad (4)$$

where $N$ is the number of training samples. $\alpha_j$ denotes on the task importance. We use ($\alpha_{det} = 1, \alpha_{box} = 0.5, \alpha_{landmark} = 0.5$) in P-Net and R-Net, while ($\alpha_{det} = 1, \alpha_{box} = 0.5, \alpha_{landmark} = 1$) in O-Net for more accurate facial landmarks localization. $\beta_i^j \in \{0,1\}$ is the sample type indicator. In this case, it is natural to employ stochastic gradient descent to train the CNNs.

*5) Online Hard sample mining:* Different from conducting traditional hard sample mining after original classifier had been trained, we do online hard sample mining in face classification task to be adaptive to the training process.

In particular, in each mini-batch, we sort the loss computed in the forward propagation phase from all samples and select the top 70% of them as hard samples. Then we only compute the gradient from the hard samples in the backward propagation phase. That means we ignore the easy samples that are less helpful to strengthen the detector while training. Experiments show that this strategy yields better performance without manual sample selection. Its effectiveness is demonstrated in the Section III.

## III. EXPERIMENTS

In this section, we first evaluate the effectiveness of the proposed hard sample mining strategy. Then we compare our face detector and alignment against the state-of-the-art methods in Face Detection Data Set and Benchmark (FDDB) [25], WIDER FACE [24], and Annotated Facial Landmarks in the Wild (AFLW) benchmark [8]. FDDB dataset contains the annotations for 5,171 faces in a set of 2,845 images. WIDER FACE dataset consists of 393,703 labeled face bounding boxes in 32,203 images where 50% of them for testing into three subsets according to the difficulty of images, 40% for training and the remaining for validation. AFLW contains the facial landmarks annotations for 24,386 faces and we use the same test subset as [22]. Finally, we evaluate the computational efficiency of our face detector.

### A. Training Data

Since we jointly perform face detection and alignment, here we use four different kinds of data annotation in our training process: (i) Negatives: Regions that the Intersection-over-Union (IoU) ratio less than 0.3 to any ground-truth faces; (ii) Positives: IoU above 0.65 to a ground truth face; (iii) Part faces: IoU between 0.4 and 0.65 to a ground truth face; and (iv) Landmark faces: faces labeled 5 landmarks' positions. Negatives and positives are used for face classification tasks, positives and part faces are used for bounding box regression, and landmark faces are used for facial landmark localization. The training data for each network is described as follows:

*1) P-Net:* We randomly crop several patches from WIDER FACE [24] to collect positives, negatives and part face. Then, we crop faces from CelebA [23] as landmark faces

*2) R-Net:* We use first stage of our framework to detect faces

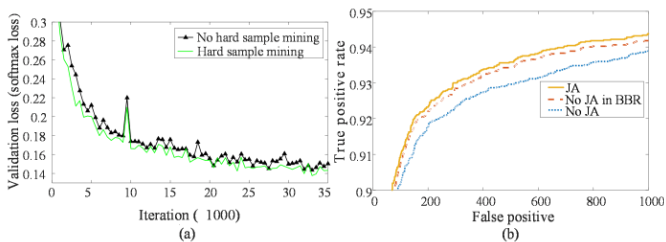

Fig. 3. (a) Validation loss of O-Net with and without hard sample mining. (b) "JA" denotes joint face alignment learning while "No JA" denotes do not joint it. "No JA in BBR" denotes do not joint it while training the CNN for bounding box regression.

from WIDER FACE [24] to collect positives, negatives and part face while landmark faces are detected from CelebA [23].

*3) O-Net:* Similar to R-Net to collect data but we use first two stages of our framework to detect faces.

### B. The effectiveness of online hard sample mining

To evaluate the contribution of the proposed online hard sample mining strategy, we train two O-Nets (with and without online hard sample mining) and compare their loss curves. To make the comparison more directly, we only train the O-Nets for the face classification task. All training parameters including the network initialization are the same in these two O-Nets. To compare them easier, we use fix learning rate. Fig. 3 (a) shows the loss curves from two different training ways. It is very clear that the hard sample mining is beneficial to performance improvement.

### C. The effectiveness of joint detection and alignment

To evaluate the contribution of joint detection and alignment, we evaluate the performances of two different O-Nets (joint facial landmarks regression task and do not joint it) on FDDB (with the same P-Net and R-Net for fair comparison). We also compare the performance of bounding box regression in these two O-Nets. Fig. 3 (b) suggests that joint landmarks localization task learning is beneficial for both face classification and bounding box regression tasks.

### D. Evaluation on face detection

To evaluate the performance of our face detection method, we compare our method against the state-of-the-art methods [1, 5, 6, 11, 18, 19, 26, 27, 28, 29] in FDDB, and the state-of-the-art methods [1, 24, 11] in WIDER FACE. Fig. 4 (a)-(d) shows that our method consistently outperforms all the previous approaches by a large margin in both the benchmarks. We also evaluate our approach on some challenge photos[1].

### E. Evaluation on face alignment

In this part, we compare the face alignment performance of our method against the following methods: RCPR [12], TSPM [7], Luxand face SDK [17], ESR [13], CDM [15], SDM [21], and TCDCN [22]. In the testing phase, there are 13 images that our method fails to detect face. So we crop the central region of these 13 images and treat them as the input for O-Net. The mean error is measured by the distances between the estimated

[1] Examples are showed in http://kpzhang93.github.io/SPL/index.html

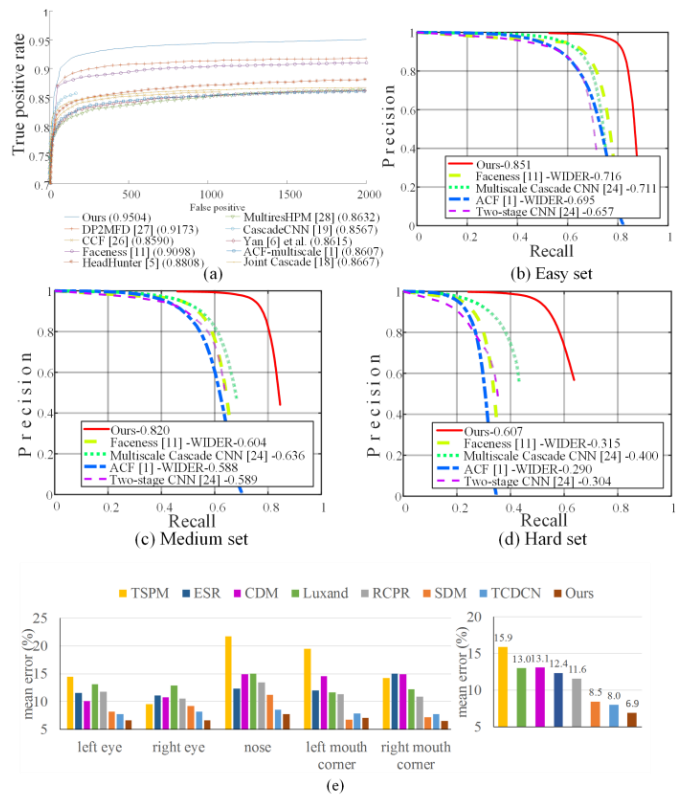

Fig. 4. (a) Evaluation on FDDB. (b-d) Evaluation on three subsets of WIDER FACE. The number following the method indicates the average accuracy. (e) Evaluation on AFLW for face alignment

landmarks and the ground truths, and normalized with respect to the inter-ocular distance. Fig. 4 (e) shows that our method outperforms all the state-of-the-art methods with a margin.

### F. Runtime efficiency

Given the cascade structure, our method can achieve very fast speed in joint face detection and alignment. It takes 16fps on a 2.60GHz CPU and 99fps on GPU (Nvidia Titan Black). Our implementation is currently based on un-optimized MATLAB code.

## IV. CONCLUSION

In this paper, we have proposed a multi-task cascaded CNNs based framework for joint face detection and alignment. Experimental results demonstrate that our methods consistently outperform the state-of-the-art methods across several challenging benchmarks (including FDDB and WIDER FACE benchmarks for face detection, and AFLW benchmark for face alignment) while keeping real time performance. In the future, we will exploit the inherent correlation between face detection and other face analysis tasks, to further improve the performance.